\documentclass[letterpaper, 10 pt, conference]{ieeeconf}  % Comment this line out if you need a4paper

\IEEEoverridecommandlockouts                              % This command is only needed if 
\usepackage{cite}
\usepackage{graphicx}
\title{\LARGE \bf
Exploratory Movement Strategies for Texture Discrimination with a Neuromorphic Tactile Sensor}

\author{Xingchen Xu$^{1,*}$, Ao Li$^{1,*}$ and Benjamin Ward-Cherrier$^{1}$,~\IEEEmembership{Member,~IEEE}% <-this % stops a space
\thanks{$^{1}$The authors are with the School of Engineering Mathematics and Technology, University of Bristol, Bristol, U.K. Corresponding Author: Benjamin Ward-Cherrier. Email:
        {\tt\small b.ward-cherrier@bristol.ac.uk}}
\thanks{*: These authors contributed equally to this work.}
\thanks{This work was partially supported by the China Scholarship Council and a Royal Academy of Engineering Fellowship on “Shared autonomy neuroprosthetics: Bridging artificial and biological touch" (RF\textbackslash202021\textbackslash20\textbackslash171).}% <-this % 
}

\begin{document}

\maketitle
\thispagestyle{empty}
\pagestyle{empty}

%%%%%%%%%%%%%%%%%%%%%%%%%%%%%%%%%%%%%%%%%%%%%%%%%%%%%%%%%%%%%%%%%%%%%%%%%%%%%%%%
\begin{abstract}
We propose a neuromorphic tactile sensing framework for robotic texture classification that is inspired by human exploratory strategies. Our system utilizes the NeuroTac sensor to capture neuromorphic tactile data during a series of exploratory motions. We first tested six distinct motions for texture classification under fixed environment: sliding, rotating, tapping, as well as the combined motions: sliding+rotating, tapping+rotating, and tapping+sliding. We chose sliding and sliding+rotating as the best motions based on final accuracy and the sample timing length needed to reach converged accuracy.
In the second experiment designed to simulate complex real-world conditions, these two motions were further evaluated under varying contact depth and speeds. Under these conditions, our framework attained the highest accuracy of 87.33\% with sliding+rotating while maintaining an extremely low power consumption of only 8.04 mW. These results suggest that the sliding+rotating motion is the optimal exploratory strategy for neuromorphic tactile sensing deployment in texture classification tasks and holds significant promise for enhancing robotic environmental interaction.

\end{abstract}

%%%%%%%%%%%%%%%%%%%%%%%%%%%%%%%%%%%%%%%%%%%%%%%%%%%%%%%%%%%%%%%%%%%%%%%%%%%%%%%%
\section{Introduction}
% 背景与动机

% 介绍研究领域的背景（例如机器人触觉感知的重要性及其在实际应用中的需求）。
% 说明该领域内存在的主要挑战或未满足的需求，激发读者兴趣。
Tactile sensing is essential for both human and robotic interaction with the environment~\cite{dahiya2009tactile}. In humans, tactile perception is inherently adaptive, with exploratory procedures (EPs) dynamically adjusted based on the properties of the object being assessed~\cite{lederman1993extracting}. Several studies have identified canonical EPs—such as lateral motion for texture, pressure for compliance, and contour following for shape—that enable efficient and targeted sensory acquisition~\cite{lederman1987hand}. In contrast, robotic tactile systems often lack this adaptability, relying on predefined and rigid exploration strategies that can limit classification performance~\cite{seminara2019active}.
% 相关工作综述

% 概述现有文献中的主要方法和技术，指出这些方法的优势和局限性。
% 强调当前研究中存在的不足或空白，明确与已有工作相比的改进空间。
% Prior research has investigated various exploratory movements for tactile object classification. Compliance-based exploration achieved through constant force application has been shown to provide reliable material discrimination, and static contact for thermal conductivity assessment proves particularly effective for rigid objects, whereas lateral sliding may suffer from limitations in force and velocity control~\cite{xu2013tactile}. Additionally, dynamic exploratory procedures that incorporate both shear and normal force measurements, such as grasping and shaking, can significantly enhance object recognition~\cite{kirby2022comparing}. Although these studies underscore the importance of effective exploratory strategies, their implementations often depend on high-power systems with fixed movement sequences, which restrict efficiency and responsiveness.
Prior research on tactile texture recognition further underscores the importance of exploratory movements. Human subjects naturally employ a variety of stereotyped actions—including lateral scanning, tapping, and rotational motions—to extract texture features effectively~\cite{callier2015kinematics}. In parallel, robotic systems have begun to explore adaptive strategies for texture discrimination. For instance, a framework utilizing the BioTac sensor combined with Bayesian exploration has demonstrated effective texture discrimination by intelligently selecting exploratory actions based on sensor measurements~\cite{fishel2012bayesian}. However, this approach was validated under ideal experimental conditions requiring precise control of contact force and sliding speed, which poses challenges for practical deployment.

% 问题陈述与研究空白

\begin{figure}[t]
    \centering
    \includegraphics[width=1\linewidth]{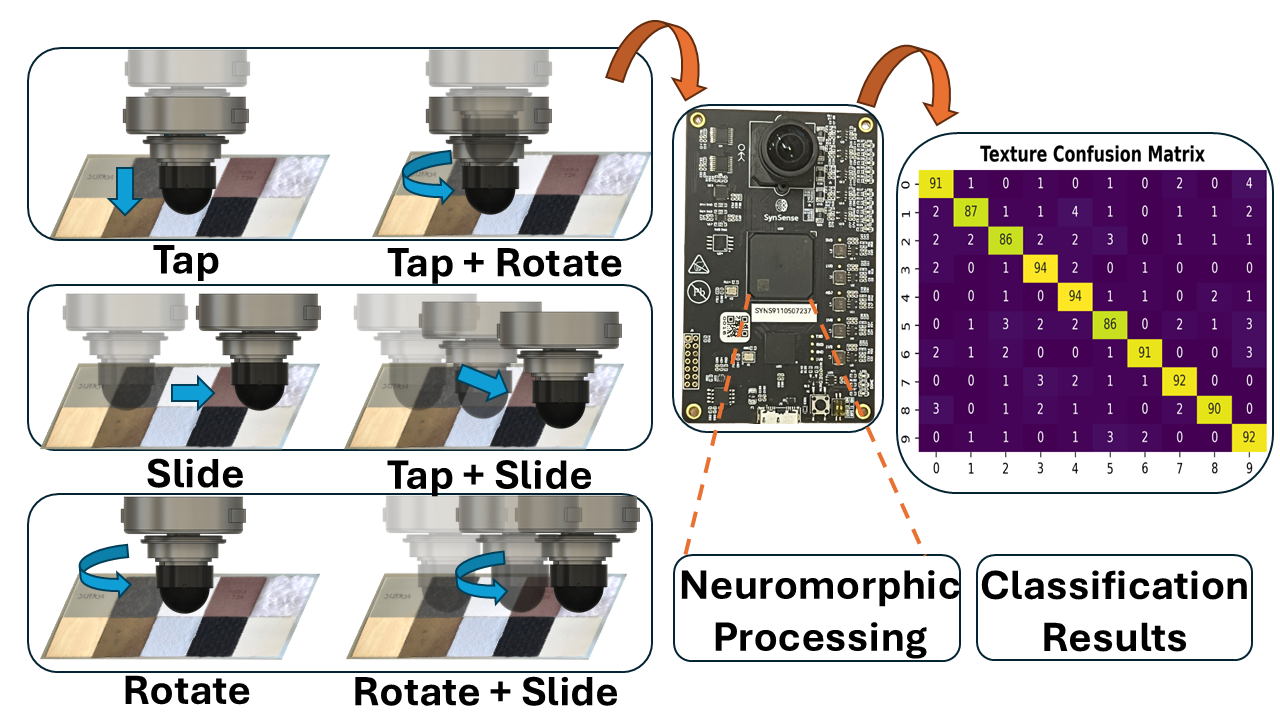}
    \caption{Overview of the experimental setup and interaction modalities. The NeuroTac sensor interacts with natural textures affixed to an acrylic platform, performing three fundamental motions: sliding, tapping, and rotating, along with their combinations. Each interaction generates neuromorphic tactile event data, which is processed and classified using a neuromorphic processor (Synsense, Speck2f) for inference.}
    \label{fig:1}
\end{figure}

% 明确指出具体问题或挑战是什么。
% 讨论现有方法无法解决的关键问题，并阐明为什么需要新的方法（例如：高功耗、反应延迟或预定运动模式的局限）
In this work, we extend these findings by adopting a neuromorphic approach for texture discrimination that is both energy-efficient and capable of rapid inference~\cite{basu2018low} (Fig.~\ref{fig:1}). Despite notable advances, existing methods for tactile texture classification often rely on computationally intensive algorithms~\cite{huang2021texture}, offline processing~\cite{brayshaw2022temporal}, and fixed exploratory motions~\cite{xie2019human} that fail to capture the inherent variability of active touch in real-world scenarios. High power consumption and delayed reaction times further constrain their applicability in resource-limited settings such as robotics and prosthetics~\cite{bartolozzi2022embodied}. To address these challenges, our approach systematically evaluates a range of human-inspired exploratory gestures (tapping, sliding, rotating, and their compound combinations) to identify the optimal strategy for tactile texture recognition.

% 主要贡献

% 用简洁的条目列出论文的主要贡献点，突出创新之处（如端到端框架、低功耗与高精度实时推断、不同手势的比较分析等）。
% \subsection{Contributions}
The main contributions of this paper are as follows:
\begin{itemize}
    \item We propose a neuromorphic tactile texture classification approach for low-power sensory inference on neuromorphic hardware.
    \item We systematically evaluate 6 exploratory motions to determine the most effective strategy for neuromorphic tactile texture recognition.
    \item We demonstrate generalisation under varying contact depths and sliding speeds, highlighting the proper depth and speed for classifying tasks.

\end{itemize}

% 论文结构概述(可选)

% 简要介绍后续章节的安排，为读者提供整体逻辑脉络。
% Our primary objective is to ascertain which interaction method—tapping, sliding, or rotating—is most effective for texture recognition. The temporal characteristics and recognition performance of each modality are discussed in detail in the Results section.
% This work contributes to the broader effort of developing \textbf{biomimetic, neuromorphic tactile systems} that enhance robotic perception and interaction with unstructured environments.

\section{Related Works}

Neuromorphic computing has emerged as a promising paradigm for efficient sensory processing in resource-constrained environments such as robotics and prosthetics~\cite{bartolozzi2022embodied}. Recent advances in spiking neural networks (SNNs) have demonstrated that these systems can perform classification tasks with significantly lower energy consumption than conventional deep learning approaches~\cite{deng2020rethinking}. Hardware platforms such as Loihi, SpiNNaker and SynSense's Speck have been successfully deployed for a variety of classification tasks~\cite{buettner2021heartbeat}%\cite{patino2020event}
\cite{dominguez2016multilayer}\cite{xu2025neuromorphic}.

Several studies have applied neuromorphic approaches specifically to tactile sensing by leveraging SNNs to process high-dimensional, event-based data from tactile sensors. For example, Patino et al.~\cite{patino2020event} proposed an event-driven framework for texture classification, while Dominguez-Morales et al.~\cite{dominguez2016multilayer} implemented an SNN-based model for object recognition using bio-inspired tactile data. Implementing these SNN models on neuromorphic processors can lead to significant decreases in power usage and inference time while maintaining high accuracy on tasks such as texture discrimination~\cite{brayshaw2024neuromorphic}. 

In parallel, various exploratory motions have been explored for texture classification. Tanauzov et al.~\cite{taunyazov2019towards} demonstrated that hybridizing touch and sliding—where an initial tactile estimate obtained from a simple touch is subsequently refined by sliding—can achieve up to 98\% classification accuracy on unseen data. Similarly, Lima et al.~\cite{lima2020dynamic} showed that dynamic exploration, particularly when incorporating sensor angular velocity, yields robust texture discrimination with accuracies reaching 95\%. Furthermore, Gupta et al.~\cite{gupta2018neuromorphic} advanced tactile classification by combining event-based tactile data with rotation-invariant feature extraction on high-contrast images, thereby mitigating the effects of variable exploratory angles and achieving average accuracies of 98\% across diverse textures and palpation directions. Collectively, these studies underscore the importance of optimizing exploratory motions and feature extraction strategies for enhanced classification performance.

Despite these advances, most research has focused on achieving robust classification under idealized conditions, while real-world tactile interactions are inherently active and variable~\cite{brayshaw2024simultaneous}. Xu et al.~\cite{xu2013tactile} have identified a set of exploratory gestures(pressure, lateral
sliding, and static contact) suitable for robotic applications, whereas Kirby et al.~\cite{kirby2022comparing} demonstrated the benefits of employing compound actions for improved object recognition. These findings suggest that bridging the gap between controlled experiments and practical deployment requires further development of gesture-specific recognition patterns that more closely mimic human touch.

Motivated by these insights, our work focuses on active tactile exploration via a neuromorphic approach that enables dynamic sensing and low-power deployment. We systematically evaluate human-inspired exploratory procedures—including tapping, sliding (lateral scanning), rotating, and their compound combinations—to identify optimal strategies for tactile texture recognition. This approach is motivated by observations in~\cite{callier2015kinematics}, where human subjects were seen to employ various exploratory movements when interacting with textures.  By leveraging the energy efficiency of neuromorphic hardware and the adaptability of event-driven processing, our framework achieves rapid, accurate classification while significantly reducing power consumption, thereby paving the way for practical deployment in robotics and prosthetic applications.

\section{Methodology}

\begin{figure*}
    \centering
    \includegraphics[width=1\linewidth]{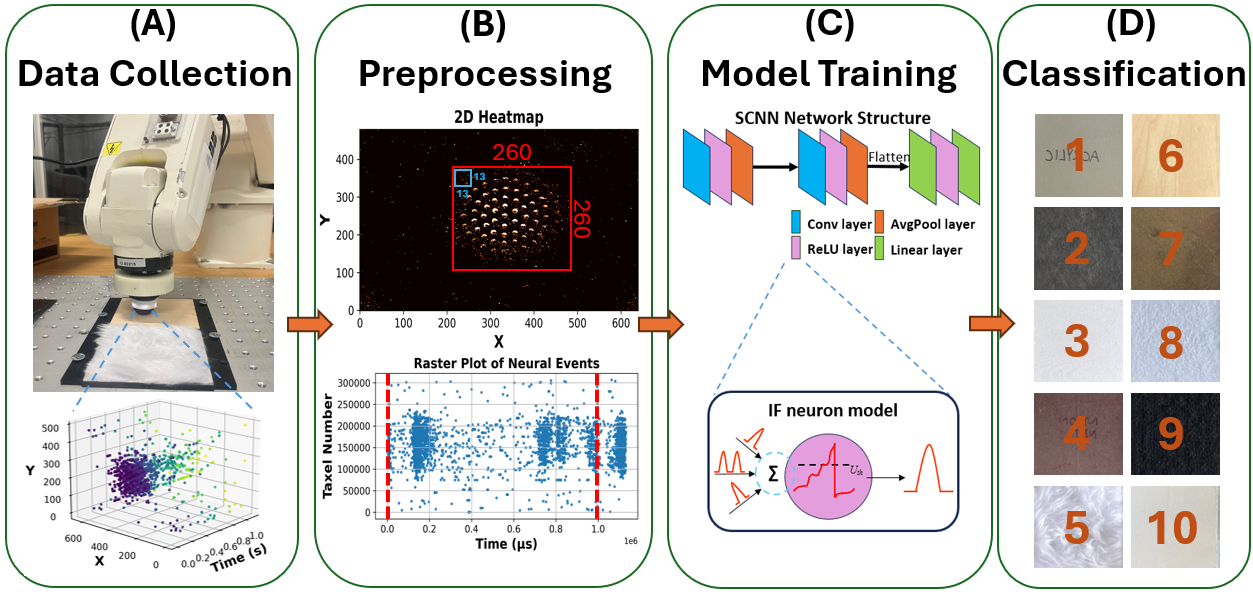}
    \caption{The pipeline of the end-to-end system. (A) The NeuroTac mounted on the robot arm (ABB, IRB120) performs one of 6 different motions on each of the 10 textures. The collected AER (Address Event Representation) data is visualized as a 3D point cloud, where the x–y plane represents the sensor's spatial resolution and the time axis captures the temporal dynamics. (B) Data preprocessing. The raw data are first cropped and pooled in spatial space (top), then clipped to 1 second in time (bottom). (C) Our spiking convolutional neural network (SCNN) model structure uses the integrate-and-fire (IF) neuron models to substitute the ReLU layer. (D) The predicted result is one of the following 10 different natural textures: 1. Acrylic, 2. Fashion Fabric, 3. Cotton, 4. Nylon, 5. Fur, 6. Wood, 7. Mesh, 8. Felt, 9. Wool, 10. Canvas.}
    \label{fig:2}
\end{figure*}

\subsection{Experiment Setup}

We utilized a NeuroTac sensor designed to replicate the rapid adaptation of FA1 afferent during dynamic contact~\cite{ward2020neurotac} to acquire tactile data. This neuromorphic vision-based sensor incorporates an event-based camera (iniVation, DVXPlorer) and features 91 internal 3D-printed white markers on a flexible black membrane. The sensor has a diameter of 20 mm, and the event-based camera outputs data in Address Event Representation (AER) format, which is directly compatible with processing via SNNs.
% \begin{figure}[htbp]
%     \centering
%     \includegraphics[width=1\linewidth]{Figs/Screenshot 2025-02-24 191024.png}
%     \caption{Real Robot}
%     \label{fig:enter-label}
% \end{figure}

The sensor was mounted on a 6-DOF robot arm (ABB, IRB120) for data collection (Fig.~\ref{fig:2}, A). 10 natural textures were trimmed to dimensions of 100\,mm by 100\,mm and then adhered to an acrylic backing using double-sided tape. These panels were subsequently placed into a groove located on the  platform beneath the robot arm. To emulate the diversity of human tactile interactions and identify the most effective modality for robotic texture classification, the robot executed six interaction motions: sliding, tapping, rotating, simultaneous sliding and tapping, simultaneous sliding and rotating, and simultaneous tapping and rotating.

% To acquire our dataset, we utilized an experimental setup featuring the neuroTac tactile sensor mounted on an ABB robotic arm, see Fig.~2. In order to mimic the diversity of human tactile interactions, the sensor was employed to execute various interaction modalities, including sliding, tapping, rotation, sliding+tapping, sliding+rotation, and tapping+rotation. Note that the duration of each interaction varies by modality, resulting in different temporal scales that serve as an additional evaluative metric in our study. 

\subsection{Data Collection}
In the first experiment, we gathered six datasets comprising three basic actions (tapping, sliding, rotating) and three compound actions (tapping+sliding, tapping+rotating, sliding+rotating). For each dataset, a single motion was applied to the 10 textures, with 100 trials per texture, yielding a total of 1,000 trials per motion type. In the case of compound actions, both motion components were performed simultaneously (e.g., during tapping+rotating, the tapping and rotating actions occurred concurrently). Each motion was executed for a duration of 1 second per trial. The detailed motion parameters are presented below:

\begin{itemize}
    \item sensor contact depth: 1.5\,mm
    \item sliding distance: 30\,mm
    \item rotation angle: 30\textdegree{}
    \item tapping speed: 1.5\,mm/s
    \item tapping and sliding speed: 30\,mm/s
    \item rotation angular speed: 30\textdegree{}/s
\end{itemize}

% the neuroTac sensor interacted with 10 different naturalistic textures using six gesture modalities derived from human behavior. These modalities comprised three basic actions---tapping, sliding, and rotating---and three compound actions---tapping~+~sliding, tapping~+~rotating, and sliding~+~rotating. It is important to note that for all gestures except tapping, the compound gestures were executed as fused motions within a 1-s interval (e.g., in tapping~+~rotating, both the tapping and rotating components were performed simultaneously). For each gesture modality, 100 trials were conducted, resulting in a total of 1,000 samples per modality. A fixed sensor contact depth of 1.5\,mm was maintained throughout these experiments. Tapping and sliding were executed at a linear speed of 30\,mm/s, with each sliding action covering a distance of 30\,mm in approximately 1\,s. Similarly, the rotational gesture was controlled such that a 30\textdegree{} rotation was completed at an angular speed of 30\textdegree{}/s in roughly 1\,s, while tapping interactions typically lasted around 0.15\,s.

In the second experiment, we first evaluated the performance of the classification system on the previously gathered 6 datasets and selected the 2 motions that demonstrated the most stable performance---the fastest 2 interactions reaching over 95\% accuracy (reason specified in \ref{result_A}) (sliding and sliding+rotation). We then assessed these best-performing motions in a generalization experiment designed to mimic real-world complexity. In the dataset, contact depth was varied between 0.5-2.5\,mm (noting that depths exceeding 3\,mm risk damaging the sensor's silicone tip). Sliding and tapping speeds were varied between 10-50\,mm/s. The angular speed was varied between 10-50\,\textdegree{}/s. A random initial position on the board---within the available motion space---was also selected for each trial. We randomly sampled locations and speeds from the described ranges with a uniform distribution. We collected 25,000 samples for the 10 textures with the 2 chosen motions, respectively. Analogously, we constrained each trial to a 1\,s duration. Our aim was to compare the generalization capabilities of the sliding and sliding+rotation interactions as well as explore the effect of depth and speed on texture classification performance.

\subsection{Preprocessing and SNN model}

The event-based camera has a resolution of \(640\times 480\). To extract valid events and decrease computation, we cropped raw sensor data into an area containing \(260 \times 260\) pixels (Fig.~\ref{fig:2}, B). This cropped area was then further divided into a grid of \(20 \times 20\) square areas, each area containing \(13 \times 13\) pixels. We aggregated individual spike trains in each small area to form a single composite spike train representing the spatiotemporal feature vector for that area. We chose 1000\,ms as the standard length for all interacting methods. Therefore, the input data after preprocessing were structured as a matrix of dimensions (1000, 20, 20), representing 1000 time steps across a \(20 \times 20\) spatial grid.

We applied spiking convolutional neural network to process the data. Our network contains two convolutional layers, two pooling layers, and three linear layers (Fig.~\ref{fig:2}, C). We used the integrate-and-fire neuron model as the substitute activation function for computation simplicity. The network was implemented in Synsense's Sinabs framework to allow for on-board processing on the Speck2f (Synsense) neuromorphic chip.

\subsection{Power Consumption Metrics}

To realize a fully integrated, end-to-end neuromorphic system with low power consumption, our SCNN model was deployed on the Synsense Speck2f neuromorphic processor (Fig.~\ref{fig:1}). Power consumption on-board was compared to a CPU (AMD Ryzen Threadripper 2920X) and a GPU (NVIDIA RTX A4500) implementation, both while idling and during inference. 

Additionally, the power consumption implications of each type of exploratory movement were explored by contrasting their power consumption during inference with the Speck2f board. 

% To compare the energy performance of our system we
% measured its power consumption when deployed across
% three different hardware configurations; onboard a Loihi2, an
% AMD Ryzen Threadripper 2920X CPU running a simulated
% Loihi2 environment and an NVIDIA RTX A4500 GPU
% running inference in PyTorch

% 5.1 纹理识别结果Acc(1000)

% 对比不同方法（CNN vs SNN）
% 在不同纹理数据集上的分类性能
% 5.2 计算效率分析

% 5.3 泛化识别分析(24000)
% 不同采样频率、滑动速度、触摸力的影响
% SNN 参数调整对分类性能的影响

% 5.4 局限性与改进方向
% 运行速度对比（GPU vs Neuromorphic Chip）
% 低功耗分析（energy efficiency）√

\section{Results}

\subsection{Performance and Evaluation of the Exploratory Motions}
\label{result_A}

\begin{figure}[t]
    \centering
    \includegraphics[width=1\linewidth]{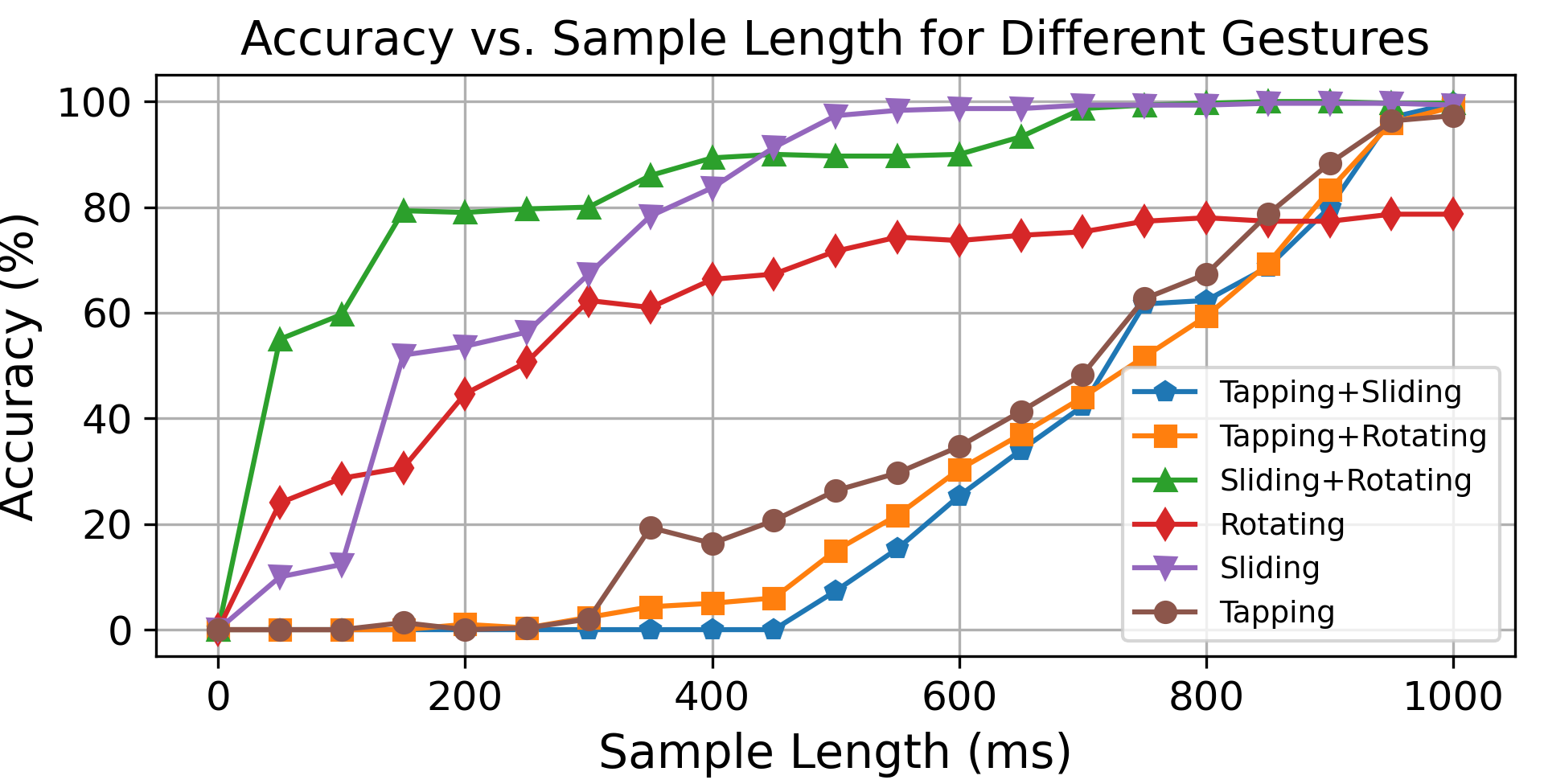}
    \caption{Inference accuracy across 6 exploratory gestures over time. The X-axis features the sample length fed into the SCNN. The total length of each sample is 1\,s. Longer sample timing length means more information for the model. Each marker gives the mean accuracy over 10 independent runs; error bars are omitted for clarity.}
    \label{fig:3}
\end{figure}

For each of the 6 exploratory motions, a single SCNN was trained. To assess the rapidity and accuracy of each motion, we segmented the data into 50 ms intervals ranging from 50\,ms to 1000\,ms.

The results indicate that classification accuracy generally improves with increasing sample length across all motions, though at varying rates (Fig:~\ref{fig:3}). Among the individual motions, sliding and sliding+rotating exhibit the highest classification performance over most sample lengths, approaching 99.5\% at the maximum duration while maintaining a faster convergence rate compared to other exploratory motions. Rotating initially achieves a relatively high accuracy but plateaus at lower final performance, indicating that while rotational motion may provide some discriminative cues early in exploration, it is insufficient to provide enough information for effective classification over the longer term. 

The results for tapping-based motions indicate the presence of a depth threshold reached at approximately 400\,ms, beyond which the model can gradually distinguish between different textures, with comparable performance between tapping, tapping+sliding and tapping+rotating. 
The tapping+rotating surpasses pure rotating at approximately 900\,ms, indicating that additional tactile information is gained when depth is varied during contact.

These findings highlight two key trends. First, sliding-based motions are particularly effective for texture classification, consistent with human tactile perception studies~\cite{hollins2000evidence}. Second, integrating multiple exploratory gestures may improve the inference efficiency by reaching high accuracy with a shorter sample time length, which shows that combining motion types can aid in gaining texture information. Additionally, longer sample durations yield more discriminative features, further improving recognition accuracy across all conditions. These insights provide a foundation for selecting optimal exploratory gestures in neuromorphic tactile perception systems. Final accuracy and sample timing length required to reach a 5\% error band of the final accuracy are two key metrics for evaluating and optimizing the motion's performance.

\subsection{Power Requirements for the Exploratory Motions}

\begin{figure}[htbp]
    \centering
    \includegraphics[width=1\linewidth]{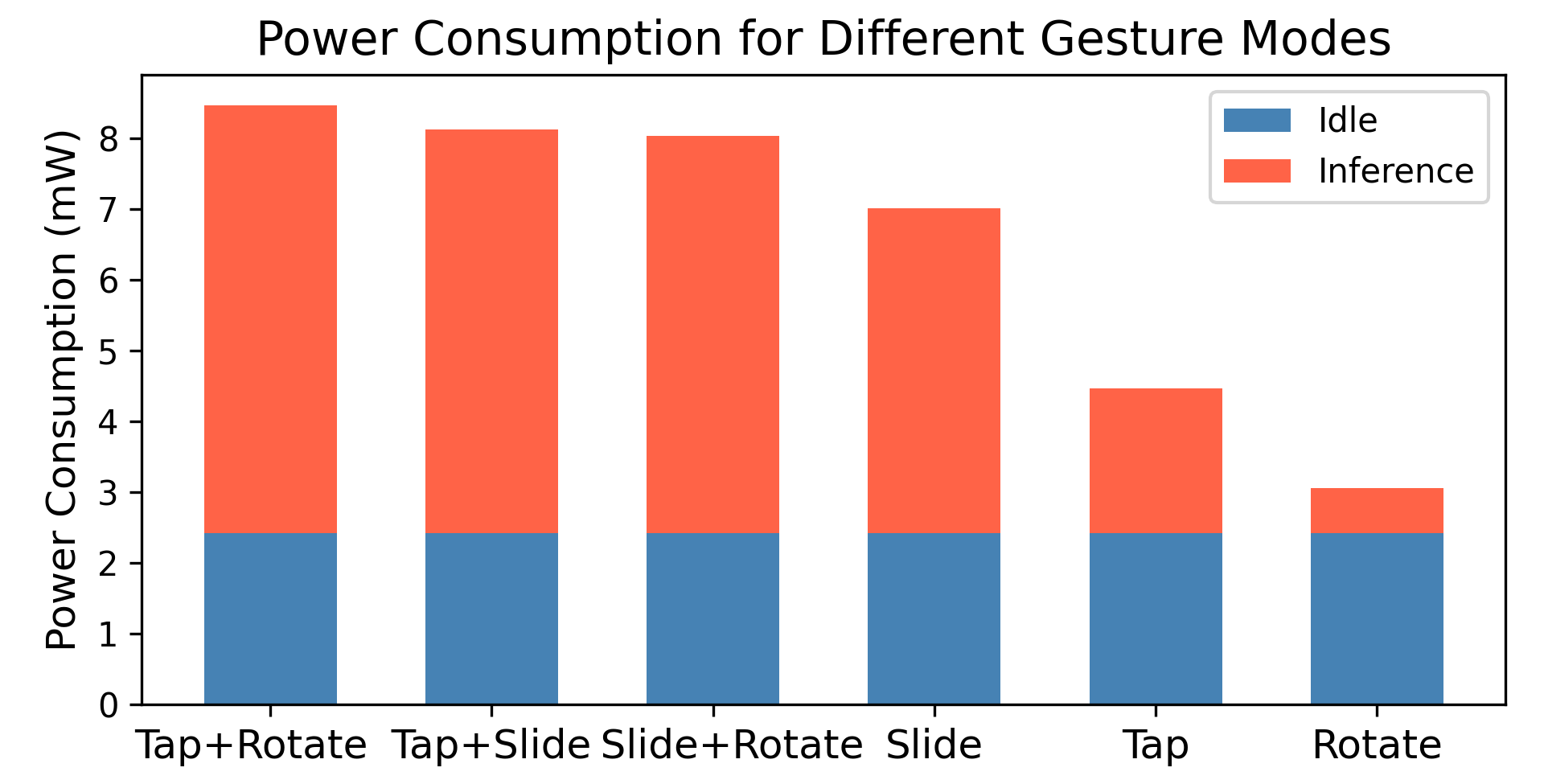}
    \caption{Power consumption across different exploratory gestures. Power consumption during inference of six exploratory motions ordered from highest to lowest power usage, measured in milliWatts. A higher event rate per unit of time results in increased average power consumption. Stacked bars differentiate idling power from additional inference power.}
    \label{fig:power}
\end{figure}

To assess the energy efficiency of our neuromorphic framework, we compared the average power consumption of tactile classification tasks executed on a conventional CPU and GPU against inference performed on the neuromorphic computing board. We first present the average power consumption on the CPU, GPU and the neuromorphic device across all motions (Table:~\ref{table:power}) then display the power required by different motion types (Fig:~\ref{fig:power}).

The average power consumption of the Speck2f neuromorphic chip is 6.53\,mW derived by the on-chip power monitor, which is over 10,000 times lower than that of CPU-based inference (67\,W) and about 2,500 times lower than that of GPU-based inference (16.31\,W). This substantial reduction demonstrates the remarkable energy efficiency of SNN-based architectures operating on neuromorphic edge devices. Among the exploratory gestures, compound motions exhibit higher power usage, with tapping+rotating consuming 8.47\,mW, tapping+sliding 8.13\,mW, and sliding+rotating 8.04\,mW. In contrast, single motions require less power—tapping at 4.47\,mW, rotating at 3.06\,mW, and sliding at 7.01\,mW. 
Although compound motions such as sliding+rotating require slightly more power than simpler gestures, the incremental energy cost on neuromorphic systems is minimal compared to traditional CPU or GPU-based inference. This negligible difference in power consumption supports the feasibility of incorporating more complex exploratory movements into the tactile sensing framework.

\begin{table}
    \centering
    \caption{Average Power Consumption on Different Devices}
    \begin{tabular}{|c|c|c|}
        \hline
        \textbf{} & \textbf{idling} & \textbf{operation}\\
        \hline
        CPU & 53.75\,W & 67.00\,W \\
        \hline
        GPU & 15.95\,W & 16.31\,W \\
        \hline
        Speck2f & 2.42\,mW & 6.53\,mW \\
        \hline
    \end{tabular}
    \label{table:power}
\end{table}

\subsection{Generalization across Different Dimensions}

Based on the inference accuracy presented in Fig.~\ref{fig:3}, we choose sliding and sliding+rotating as the 2 most accurate exploratory motion types on the texture classification task (the first two to reach 95\% accuracy).

To mimic the real-world scenarios where dimensions like speed and depth of contact play an important role in texture perception~\cite{brayshaw2024simultaneous}, we conducted experiments over a range of varying conditions including depth, sliding speed and angular speed.% In these experiments, contact depth was varied from 0.5\,mm to 2.5\,mm and sliding speed from 10\,mm/s to 50\,mm/s with randomized values. %For each combination, classification accuracy was recorded, yielding a comprehensive dataset that elucidates the dependency of classification performance on these operational parameters.

% Among the two selected motions, sliding+rotating achieves the highest converged accuracy in the generalized model when tested on unseen data across linear speed (10-50\,mm/s), angular speed (10-50\textdegree{}/s), and depth (0.5-2.5\,mm) (Table~\ref{table:general_models}), while the sliding motion is evaluated over linear speed and depth within the same range.

Among the two selected motions, sliding + rotating converges to an accuracy of 87.33\,\%, compared with 79.67\,\% for pure sliding, when the generalized model is tested on unseen data across linear speeds of 10-50\,mm/s, angular speeds of 10-50\textdegree{}/s, and indentation depths of 0.5-2.5\,mm. This further supports the idea that compound motions can enhance texture recognition by incorporating additional tactile information. Notably, in human texture discrimination, lateral sliding is often accompanied by subtle rotational movements~\cite{lezkan2018interdependences}, reinforcing the biological plausibility of this approach.

\begin{figure}[htbp]
    \centering
    \includegraphics[width=1\linewidth]{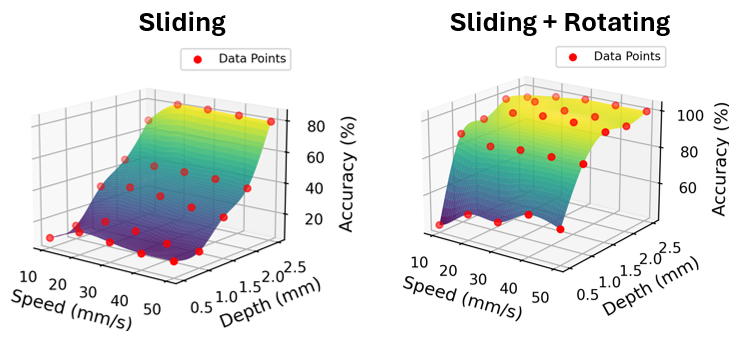}
    \caption{Classification accuracy of the generalized model for different sliding speeds and contact depths. %(the angular speed is numerically equivalent to the linear speed in value). 
    Red dots show the accuracy of the model on the specified combination of speed and depth.}
    \label{fig:wave}
    % \vspace{-8pt}
\end{figure}

We also recorded the performance of the general models under different combinations of speed (linear speed: 10-50\,mm/s, angular speed: 10-50\textdegree{}/s) and depth (0.5-2.5\,mm) (Fig.~\ref{fig:wave}).

Variations in sliding speed produced only moderate effects on classification accuracy. Only the slowest speed (10\,mm/s) produced a slight decrease in accuracy, and this effect was slightly more pronounced at shallower depths. The system maintained robust performance across the rest of the speed range (20–50\,mm/s).  %Notably, an intermediate speed (approximately 30\,mm/s) appears to provide a favorable balance between performance and practical motion execution.
Contact depth however appears to play a critical role in classification accuracy, with greater depths performing best. A greater depth results in closer contact between the sensor and the texture, allowing for a larger area of the sensor's skin contacting with the texture and increased frictional forces. Increased depth therefore generates more events, providing richer sensory data for improved texture recognition. % At a shallow contact of 0.5\,mm, accuracies ranged from 9.33\% to 17.33\%, suggesting that such minimal contact fails to provide sufficient tactile cues for robust texture discrimination. Increasing the contact depth to 1.0\,mm resulted in only modest improvements (11.67\%–18.67\%), whereas a further increase to 1.5\,mm yielded a significant boost, with accuracies rising to approximately 29\%–35\%. At 2.0\,mm, the performance improved further to around 42\%–45\%, and the highest accuracies—ranging from 67.33\% to 81.00\%—were observed at 2.5\,mm.
Comparing the left and right panels of Fig.~\ref{fig:wave} also suggests that incorporating rotation in the exploratory motion significantly expands the range of depths which produces highly accurate classification (above 80\%) from only 2.5 mm to 1–2.5 mm. This highlights the crucial role of rotation in texture discrimination, particularly in real-world scenarios where generalization across varying conditions is essential.

% \begin{table}
%     \centering
%     \caption{Accuracy of the 2 motions on their generalized models}
%     \begin{tabular}{|c|c|c|}
%         \hline
%         \textbf{} & \textbf{sliding} & \textbf{sliding+rotating}\\
%         \hline
%         accuracy & 79.67\% & 87.33\%\\
%         \hline
%     \end{tabular}
%     \label{table:general_models}
% \end{table}

These findings suggest that for effective generalized texture classification, contact depth is a key parameter of the exploratory motion while rotating the sensor aids the accuracy, convergence speed and generalization of the neuromorphic tactile system.

\section{Discussion}
% Result p1: 和人类相关的动作选择
In this study,  we applied a low-power, neuromorphic tactile sensing system to a texture classification task with 6 different exploratory motions (tapping, sliding, rotating, tapping+rotating, tapping+sliding and sliding+rotating). We initially examined the performance of the exploratory motions by analyzing their classification accuracy over increasing sample timing lengths. The results indicate that while most motions achieve satisfactory accuracy over the full sample time of 1000\,ms, sliding and sliding+rotating outperform other motions in both final accuracy and the minimal sample timing length required to reach a 5\% error band of the final convergence accuracy. We investigated these motions further and demonstrated the importance of adding rotation to the sliding motion in generalizing to different depths. This observation may be linked to the complex contact mechanics at the interface between the human fingertip and explored surfaces~\cite{delhaye2014dynamics}, particularly given that our NeuroTac sensor is designed to emulate FA1 receptor function. In related robotic texture classification experiments, the angular rotating~\cite{lima2020dynamic} and linear sliding motions\cite{xie2019human} have been proven to enable rapid and precise texture classification.  These findings are corroborated by observations of human tactile exploration, where lateral scanning and compound motions are routinely employed to swiftly and accurately discern texture characteristics\cite{fishel2012bayesian}~\cite{fulkerson2013first}.

% Result p2: 功耗设计
The deployment of the neuromorphic board highlights the advancements of SNNs and neuromorphic hardware in energy-efficient computation. The system's extreme low power consumption (below 9\,mW) is significantly lower than comparable neuromorphic devices, such as Loihi (31\,mW~\cite{muller2022braille}). The number of events per unit time directly influences power consumption on the board, explaining why rotation consumes the least power, as it generates the fewest events. However, across all motion types, the on-board power consumption remains significantly lower compared to CPU and GPU implementations. This efficiency makes edge computing feasible for integration into prosthetics and other real-time, power-constrained tactile applications~\cite{han2022review}.
% Result p3: 真实世界的深度和速度选择，以及准确率讨论
%I would delete this and replace with a paragraph discussing how these kinds of exploratory motions could exist for all sorts of tactile interactions going beyond textures  and link to the literature on exploratory procedures, and anything suggesting this is the case for artificial or human touch

Moreover, recent studies emphasize the value of exploring motions (tapping and pressing) for capturing tactile properties beyond texture discrimination, such as shape, compliance, and thermal conductivity~\cite{bok2021texture}. In human touch, strategies like lateral scanning and pressure modulation are key for extracting rich tactile cues~\cite{lederman1987hand}. Dynamic exploratory procedures in artificial tactile systems likewise boost performance; for example, Kirby et al. report a 38\% increase in recognition accuracy over static, single-touch methods \cite{kirby2022comparing}. Extending Rongala et al.’s spike-based approach~\cite{rongala2015neuromorphic}, we introduce more intricate motion patterns and shorter sensing windows. These findings support the integration of diverse exploratory motions into robotic systems to enhance tactile perception across multiple dimensions.

In future work, we plan to extend our investigations to more complex motion patterns and environmental scenarios, incorporating variable contact angles and a sensorimotor control loop for active touch~\cite{martinez2017active} that modulates contact force and sliding speed within a single trial. Such extensions aim to better mimic the variability of natural human tactile exploration~\cite{callier2015kinematics} and potentially enable real-time adjustments based on sensory feedback.

Our goal is to develop a robust and adaptive tactile sensing framework that delivers high-precision texture classification under real-world conditions. We intend to systematically evaluate the proposed method across a range of advanced tactile sensors, including Gelsight~\cite{yuan2017gelsight} and Evetac\cite{funk2024evetac}, to further establish its generalizability and effectiveness in artificial tactile sensing.

\section{Conclusion}
In this work, we applied an end-to-end neuromorphic tactile system to texture classification and demonstrated that sliding-based motions serve as a highly effective exploratory strategy for this task. Additionally, we highlighted the importance of adding rotation to the sliding motion to enhance accuracy (from 79.7\% to 87.3\%) and generalization across depths. We showed that compound motions such as sliding+rotating have a minimal effect on power consumption for neuromorphic systems, opening the door for more complex exploratory motions to be investigated for texture classification and other tactile tasks.

% \addtolength{\textheight}{-12cm}   



% \section*{ACKNOWLEDGMENT}

%%%%%%%%%%%%%%%%%%%%%%%%%%%%%%%%%%%%%%%%%%%%%%%%%%%%%%%%%%%%%%%%%%%%%%%%%%%%%%%%

\bibliography{Reference}

\end{document}